\pdfoutput=1
\documentclass[runningheads,a4paper]{llncs}

\usepackage[T1]{fontenc}
\usepackage{graphicx}
\usepackage{booktabs}
\usepackage{amsmath}
\usepackage{amssymb}
\usepackage{multirow}
\usepackage{color}
\usepackage{url}

\usepackage[colorlinks=true,urlcolor=blue,linkcolor=blue,citecolor=blue]{hyperref}
\makeatletter
\g@addto@macro\UrlBreaks{\do\_\do\-}
\makeatother

\begin{document}

\title{Predictive Entropy as a Joint Screen for Error and Paraphrase Instability in Medical Vision-Language Models}
\titlerunning{Entropy as a Joint Error and Paraphrase Screen}

\author{Binesh Sadanandan\and Vahid Behzadan}
\authorrunning{B. Sadanandan and V. Behzadan}
\institute{SAIL Lab, University of New Haven, West Haven, CT, USA\\
\email{\{bsada1@unh,vbehzadan@\}newhaven.edu}}

\maketitle

\begin{abstract}
Medical Vision-Language Models (VLMs) answering binary presence questions on chest radiographs can fail in two linked ways: they are confidently wrong, and they change answers when a clinically equivalent question is rephrased. In a binary answer head both failures track the same logit margin, so we test how well one score screens for both. On MedGemma-4B-IT across MIMIC-CXR (in-distribution, $n{=}196$) and PadChest (out-of-distribution, $n{=}861$), single-pass predictive entropy predicts which yes/no answers flip under rephrasing with an area under the receiver operating characteristic curve (AUROC) of 0.823 on PadChest and 0.821 on MIMIC, and the signal is stable across five random seeds ($0.828\pm0.021$). The same signal replicates on LLaVA-RAD-7B (AUROC 0.83) and is invariant to the flip definition (all, operator-preserving, or negation-excluded rewrites), so the screen needs no semantic filter at inference. One low-entropy threshold drives error and flip rate down together (3.3\% and 5.7\% at 80\% coverage), and threshold selection transfers to a held-out split. Under cross-site shift the more expensive methods we test add nothing practical: at the 5\% risk target a single forward pass answers 88.5\% of cases, matching Monte Carlo (MC) Dropout, and beats the uniformly averaged five-seed adapter ensemble on error detection by 0.111 AUROC. Even a model that is 91\% accurate still contradicts itself on 13\% of items under rephrasing, and one forward pass flags them.
\end{abstract}

\section{Introduction}\label{sec:intro}

A radiologist asks a Vision-Language Model (VLM) whether a chest X-ray shows pleural effusion. The model answers ``Yes'' with 94\% softmax confidence, and the answer is wrong. Its confidence gives no warning. Ask the same question in slightly different words and the answer can flip to ``No.'' In medical AI, miscalibrated confidence carries a direct patient-safety cost: when clinicians cannot tell reliable predictions from confident errors, they lose the cue to seek a second opinion or escalate an ambiguous case~\cite{kompa2021second,jiang2012calibrating}. Safe clinical use of medical VLMs~\cite{medgemma2025} therefore depends on a property most evaluations skip: well-calibrated uncertainty.

Three problems make uncertainty quantification (UQ) urgent here. Softmax probabilities from modern networks are overconfident~\cite{guo2017calibration}, and fine-tuning can worsen the overconfidence. Clinical deployment brings distribution shift through new scanners, demographics, and acquisition quality, and calibration that looks fine in-distribution can break under mild shift~\cite{ovadia2019can}. Medical VLMs also have a separate fragility: they give contradictory answers when a question is reworded with no change in meaning~\cite{sadanandan2026psfmed}. Prior work treats miscalibration and paraphrase sensitivity as independent. For a binary answer both are read off the same logit margin, so whether one score screens for both is an empirical question.

Existing medical-VQA benchmarks report accuracy and F1 but not whether confidence tracks correctness~\cite{lau2018vqarad,liu2021slake}, and recent robustness studies show VLMs degrade under imaging artifacts without measuring whether their uncertainty stays reliable~\cite{kahl2025surevqa}. Meanwhile UQ has mature tools: temperature scaling~\cite{guo2017calibration}, Monte Carlo (MC) Dropout~\cite{gal2016dropout}, deep ensembles~\cite{lakshminarayanan2017simple}, and conformal prediction~\cite{angelopoulos2021gentle}. Low-Rank Adaptation (LoRA) ensembles have been proposed as a parameter-efficient route to calibrated uncertainty~\cite{wang2023loraensembles,turkoglu2024loraensemble}, but they are typically validated in-distribution, and Bayesian deep learning can fail on subtle near-distribution shifts~\cite{seligmann2023beyond}. LoRA ensemble behavior under the cross-site shift that defines real deployment is untested.

We benchmark four uncertainty quantification methods for binary presence VQA on chest radiographs with MedGemma-4B-IT~\cite{medgemma2025} across MIMIC-CXR (in-distribution) and PadChest (out-of-distribution), with cross-architecture replication on LLaVA-RAD-7B~\cite{chaves2024llavarad}. In a binary decision predictive entropy is a monotone transform of the absolute logit margin, so a coupling between instability and small margins is expected; our contribution is not that mechanism but a measurement of how strong, stable, and deployable the resulting screen is. Our contributions:
\begin{enumerate}
\item \textbf{The link, quantified.} Single-pass predictive entropy predicts which answers flip under rephrasing at AUROC (area under the receiver operating characteristic curve) 0.823 on PadChest and 0.821 on MIMIC, stable across five adapters. One entropy threshold flags both unreliable and rephrase-sensitive answers.
\item \textbf{Invariance to the flip definition.} The link is nearly unchanged on all, operator-preserving, or negation-excluded rewrites, so the screen needs no operator classifier or semantic filter at inference.
\item \textbf{Cross-architecture replication.} The link holds on LLaVA-RAD-7B (AUROC 0.83 fine-tuned): the margin--entropy relation holds for any binary answer head, and the measured strength of the screen transfers with it.
\item \textbf{One pass suffices under shift.} Out-of-distribution, MC Dropout on the adapter and a uniformly averaged five-seed ensemble give no practical gain over one forward pass: one pass matches MC Dropout on coverage at the 5\% risk target and beats the ensemble for error detection ($+0.111$ AUROC). In-distribution the ensemble leads (Table~\ref{tab:calib}); selected, weighted, or jointly calibrated ensembles are untested.
\end{enumerate}
Even an accurate model (91\%, flip rate cut from 74\% to 13\% by targeted fine-tuning) still contradicts itself on a residual 13\% of items, and a single forward pass of entropy catches them.

\section{Background and Methods}\label{sec:methods}

\paragraph{VLM as a binary classifier.}
MedGemma-4B-IT~\cite{medgemma2025} pairs a SigLIP encoder (256 image tokens) with a 34-layer Gemma~3 backbone. Given a radiograph $\mathbf{x}$ and a binary question $q$, we read the first-token logits and keep the ``Yes'' and ``No'' entries, $z_+$ and $z_-$. The positive-class probability is $\hat p=\sigma(z_+-z_-)$, reducing the generative model to a binary classifier for calibration analysis. This Yes/No reduction is standard for binary presence VQA and matches the PSF-Med paraphrase benchmark~\cite{sadanandan2026psfmed}; free-text generation is out of scope here. Predictive entropy is $\mathbb{H}[\hat p]=-\hat p\log\hat p-(1-\hat p)\log(1-\hat p)$; in this binary setting it is a monotone function of the absolute margin $|z_+-z_-|$, so the two rank items identically. We report entropy because it is a probability-scale quantity comparable across models and extends to the multiclass and free-text settings where the margin equivalence breaks. With $S$ stochastic passes (MC Dropout or an ensemble), total entropy decomposes into an aleatoric part (mean member entropy) and an epistemic part, the mutual information $\mathbb{I}(Y;\theta\mid\mathbf{x})$~\cite{depeweg2018decomposition,smith2018understanding}.

\paragraph{UQ methods.}
We evaluate \textsc{Softmax} entropy from one forward pass; \textsc{TempScale}, a scalar $T$ fit by negative log-likelihood (NLL) on a 15\% calibration split~\cite{guo2017calibration}; \textsc{MC-Drop}, $K{=}10$ stochastic passes with the LoRA adapter dropout ($p_{\text{drop}}{=}0.05$) left active~\cite{gal2016dropout}; and a five-seed \textsc{Ensemble} of independently trained LoRA adapters~\cite{lakshminarayanan2017simple}. We add \textsc{Margin}, the absolute logit gap $|z_+-z_-|$, as a ranking ablation of \textsc{Softmax}: it shares $\hat p$ by construction, so their calibration metrics coincide and they can differ only in selective-prediction ranking. This gives five scoring variants derived from four uncertainty methods.

\begin{figure}[t]
\centering
\includegraphics[width=0.56\textwidth]{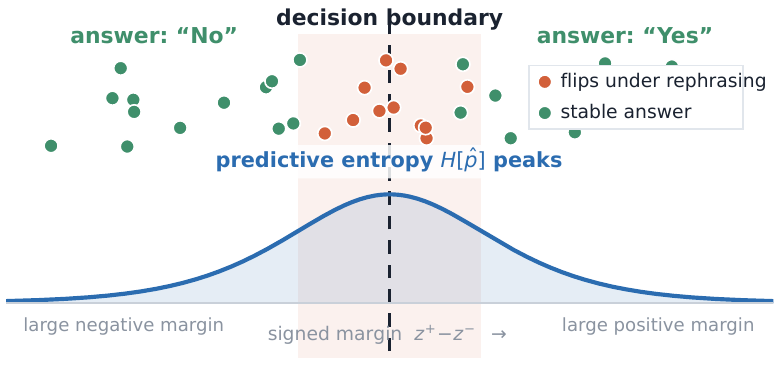}
\caption{Why one score screens both failures (schematic). A binary question sits at a signed margin $z^{+}\!-\!z^{-}$ from the decision boundary; far from it the answer stays stable under rephrasing, while near it the margin is small, predictive entropy $H[\hat p]$ peaks, and a reworded question is most likely to flip. On PadChest, flip-prone Targeted LoRA items carry about 0.17 nats more entropy than stable ones (Mann-Whitney $p<10^{-27}$), and entropy predicts flips at AUROC 0.823.}
\label{fig:geometry}
\end{figure}

\paragraph{The uncertainty--paraphrase link.}
For each item we collect predictions under its paraphrases from the PSF-Med bank~\cite{sadanandan2026psfmed}; a \emph{flip} occurs if any paraphrase changes the binary answer. The expected relation is simple. When the margin is large, rephrasing perturbs the representation but not enough to cross the boundary; when the margin is near zero, a small perturbation flips the answer, and high entropy marks exactly that near-boundary regime (Fig.~\ref{fig:geometry}). We test this link two ways: a Mann-Whitney $U$ comparison of entropy between flipped and stable items, and the AUROC of each uncertainty score as a flip predictor. Flip-prediction AUROC is a ranking statistic and must be read with accuracy and flip prevalence: a near-constant classifier flips rarely, yet ranks those few flips well while being clinically useless. Not every PSF-Med rewrite is answer-preserving. A rewrite that changes the clinical operator, for example ``Is X present?'' becoming ``Can you rule out X?'', should change a yes/no answer, so we also report an auditable \emph{operator-preserving} subset using the bank's rule-based per-rewrite annotations, and an \emph{ex-negation} subset that further removes polarity-inverting rewrites. Each rewrite's operator type is fixed metadata, assigned before any model scoring and never read from the model's outputs, uncertainty, or correctness, so these subsets test the link under a stricter definition of answer-preserving paraphrase.

\paragraph{Setup.}
The primary model is a Targeted LoRA (rank 16, $\alpha{=}32$, dropout 0.05) on layers 15--19 of MedGemma, trained 3 epochs on $n{=}2{,}000$ binary MIMIC-CXR samples with a consistency-accuracy loss~\cite{sadanandan2026psflora}; it cuts the PadChest flip rate from 74\% to 13\% while lifting accuracy, and supplies the five ensemble seeds. A Full LoRA (all 34 layers) and the base checkpoint are included for contrast. We replicate the full pipeline on LLaVA-RAD-7B~\cite{chaves2024llavarad} (Vicuna-7B + CLIP-ViT-L/14, 32 layers) with proportionally matched layers 14--18. Evaluation uses MIMIC-CXR binary presence questions ($n{=}196$, in-distribution) and the PadChest flip bank ($n{=}861$, out-of-distribution, $\approx$5 paraphrases each)~\cite{johnson2019mimic,bustos2020padchest}; the evaluation questions are disjoint from training. PadChest PNGs are 16-bit, so we percentile-window them (0.5--99.5) before the 8-bit conversion the encoder expects. We report Expected Calibration Error (ECE, 15 confidence bins), Brier, NLL, area under the risk-coverage curve (AURC), flip-prediction AUROC, and Mann-Whitney $p$-values, all with 95\% bootstrap confidence intervals ($B{=}2{,}000$). Inference runs on NVIDIA A100-80GB GPUs.

\section{Results}\label{sec:results}

\subsection{Calibration}\label{sec:res:cal}

The models are competent on out-of-distribution PadChest: the base model reaches 78.7\% accuracy and Targeted LoRA 91.4\% (Table~\ref{tab:calib}). Calibration ranks them the same way. Base ECE is 15.9\%, Full LoRA is worst at 26.9\% (here adapting all 34 layers hurts out-of-distribution calibration), and Targeted LoRA reaches 11.1\% with a low AURC (0.019). A scalar temperature is the cheapest improvement, dropping Targeted LoRA ECE to 3.6\% at the cost of one held-out fit. MC Dropout matches softmax (ECE 10.9\%, AURC 0.019). The five-seed ensemble reaches 72.6\% accuracy and does not improve over a single pass (AURC 0.13). \textsc{Margin} shares $\hat p$ with \textsc{Softmax} and coincides on every metric. In-distribution the ordering reverses (Table~\ref{tab:calib}): the ensemble leads on every metric (0.857 accuracy, AURC 0.030), still below its best single member (87.2\%; supplement), so its failure is specific to cross-site shift rather than to ensembling. Targeted LoRA likewise trades in-distribution accuracy (0.806 vs.\ 0.847) for out-of-distribution accuracy (0.914 vs.\ 0.787).

Per-member diagnostics (supplement) show the seeds vary out-of-distribution (52--92\% accuracy) while per-seed in-distribution accuracy spans 77.6--87.2\%; probability averaging does not recover the best adapter. Seed-only LoRA ensembles, averaged without member selection or weighting, thus give no benefit under this cross-site shift.

\begin{table}[t]
\centering
\caption{Calibration out-of-distribution on PadChest and in-distribution on MIMIC-CXR (questions disjoint from training, images overlapping). Lower is better except accuracy. \textsc{Margin} shares $\hat p$ with \textsc{Softmax} exactly and coincides on every metric; \textsc{MC-Drop} averages $K{=}10$ stochastic passes and ranks items slightly differently. \textsc{TempScale} is scored on the items outside its calibration split. Bold marks the best value per column within each panel. Out-of-distribution a scalar temperature gives the lowest ECE and the ensemble does not improve on one pass; in-distribution the ordering reverses. AURC is ambiguous in its third decimal where confidences tie ($\pm0.002$). Bootstrap 95\% CIs for ECE (2{,}000 resamples): Targeted \textsc{Softmax} [0.101, 0.129] out-of-distribution, [0.09, 0.16] in-distribution.}
\label{tab:calib}
{\small
\begin{tabular}{@{}lccccc@{}}
\toprule
Method & Acc $\uparrow$ & ECE $\downarrow$ & Brier $\downarrow$ & NLL $\downarrow$ & AURC $\downarrow$\\
\midrule
\multicolumn{6}{@{}l}{\emph{PadChest, out-of-distribution ($n{=}861$; 732 for \textsc{TempScale})}}\\
\midrule
Base \textsc{Softmax}        & 0.787 & 0.159 & 0.174 & 0.622 & 0.057\\
Full LoRA \textsc{Softmax}   & 0.662 & 0.269 & 0.290 & 1.153 & 0.362\\
Targeted LoRA \textsc{Softmax}   & 0.914 & 0.111 & 0.075 & 0.278 & \textbf{0.019}\\
Targeted LoRA \textsc{TempScale} & 0.910 & \textbf{0.036} & \textbf{0.068} & \textbf{0.234} & 0.022\\
Targeted LoRA \textsc{MC-Drop}   & \textbf{0.916} & 0.109 & 0.075 & 0.278 & 0.019\\
Targeted LoRA \textsc{Ensemble}  & 0.726 & 0.078 & 0.176 & 0.521 & 0.130\\
Targeted LoRA \textsc{Margin}    & 0.914 & 0.111 & 0.075 & 0.278 & \textbf{0.019}\\
\midrule
\multicolumn{6}{@{}l}{\emph{MIMIC-CXR, in-distribution ($n{=}196$; 167 for \textsc{TempScale})}}\\
\midrule
Base \textsc{Softmax}        & 0.847 & 0.145 & 0.144 & 0.824 & 0.113\\
Full LoRA \textsc{Softmax}   & 0.842 & 0.129 & 0.139 & 0.539 & 0.065\\
Targeted LoRA \textsc{Softmax}   & 0.806 & 0.109 & 0.127 & 0.399 & 0.060\\
Targeted LoRA \textsc{TempScale} & 0.820 & 0.096 & 0.120 & 0.385 & 0.055\\
Targeted LoRA \textsc{MC-Drop}   & 0.806 & 0.112 & 0.127 & 0.400 & 0.060\\
Targeted LoRA \textsc{Ensemble}  & \textbf{0.857} & \textbf{0.092} & \textbf{0.091} & \textbf{0.304} & \textbf{0.030}\\
Targeted LoRA \textsc{Margin}    & 0.806 & 0.109 & 0.127 & 0.399 & 0.060\\
\bottomrule
\end{tabular}}
\end{table}

\subsection{The Uncertainty--Paraphrase Link}\label{sec:res:bridge}

On PadChest the base model flips on 74\% of predictions; Targeted LoRA cuts this to 13\% (114/861). The base model still reaches 78.7\% accuracy, so accuracy and rephrasing stability are distinct axes: a fragile decision boundary flips under wording even when the answer is often right. Flip-prone items carry significantly higher entropy: the flip-versus-stable entropy gap is 0.17 nats (Mann-Whitney $p<10^{-27}$), and single-pass entropy predicts flips at AUROC 0.823. The signal holds in-distribution as well (0.821 on MIMIC) and is stable across the five independently-trained adapters ($0.828\pm0.021$; per-seed values in the supplement).

The link is invariant to how a flip is defined (Table~\ref{tab:bridge}). Counting only operator-preserving rewrites gives AUROC 0.827, and further removing polarity-inverting rewrites gives 0.826; both sit inside the confidence interval of the all-paraphrase value (0.823). Rewrites that change the clinical operator or invert polarity account for only ten of the 114 flips, so they barely affect the flip count for this model. The screen therefore needs no operator classifier or semantic filter at inference.

One entropy threshold lowers error and flip rates together as coverage tightens: at 80\% coverage the error rate falls from 8.6\% to 3.3\% and the flip rate from 12.1\% to 5.7\%. Tighter settings push both below 2.5\% but abstain on most cases, so we treat them as the limit of the trade-off rather than as operating points (full sweep in the supplement). Threshold selection transfers to held-out data: choosing the threshold on a random half of PadChest and applying it to the held-out half gives error 0.6\% [0.0, 1.4] and flip 1.6\% [0.0, 3.3] at 21\% coverage (1{,}000 random splits).

\begin{table}[t]
\centering
\caption{Flip-definition invariance on PadChest (Targeted LoRA, $n{=}861$ items). Rewrites is the count of operator-matching paraphrases retained; Flip$+$ is the number of items with at least one flip in the subset (prevalence in parentheses); items with no matching rewrite count as stable. AUROC has 95\% bootstrap confidence intervals (2{,}000 resamples). The three intervals overlap, so restricting the flip definition does not change the screen.}
\label{tab:bridge}
{\small
\begin{tabular}{@{}lccc@{}}
\toprule
Subset & Rewrites & Flip$+$ items & AUROC [95\% CI]\\
\midrule
All paraphrases               & 4{,}251 & 114 (13.2\%) & 0.823 [0.778, 0.864]\\
Operator-preserving           & 3{,}501 & 107 (12.4\%) & 0.827 [0.784, 0.868]\\
Operator-preserving, ex-negation & 3{,}394 & 104 (12.1\%) & 0.826 [0.780, 0.870]\\
\bottomrule
\end{tabular}}
\end{table}

\subsection{Selective Prediction and Entropy Decomposition}\label{sec:res:select}

\begin{figure}[t]
\centering
\includegraphics[width=0.52\textwidth]{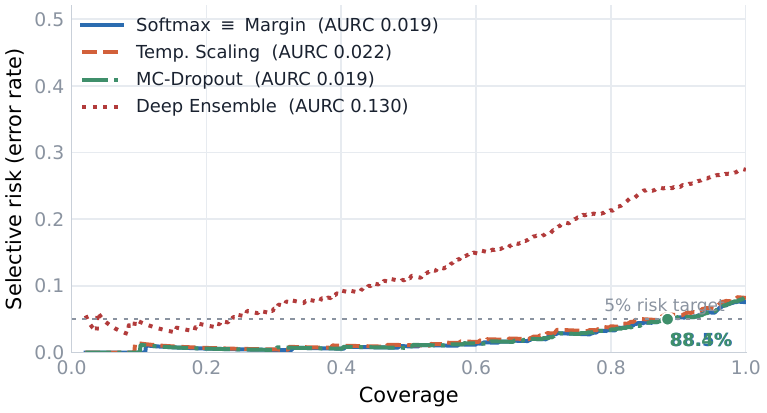}
\caption{Risk-coverage curves on PadChest (Targeted LoRA). The three single-pass methods overlap and reach the 5\% risk target (dotted line) at about 88\% coverage; the deep ensemble lags far behind (AURC 0.13).}
\label{fig:rc}
\end{figure}

At the 5\% risk target, the single-pass methods answer most of PadChest: \textsc{Softmax} reaches 88.5\% coverage, \textsc{MC-Drop} 88.4\%, and \textsc{TempScale} 86.7\%, while the ensemble reaches only 23.8\% (Fig.~\ref{fig:rc}). The three single-pass curves overlap (AURC 0.019--0.022); the ensemble is far worse (AURC 0.13). Averaging $K{=}10$ adapter-dropout passes adds nothing.

The extra passes also buy no error detection. Single-pass \textsc{Softmax} entropy gives error-detection AUROC 0.862, statistically tied with \textsc{MC-Drop} (0.856; paired difference $+0.006$ [$-0.002$, $0.017$]) and ahead of the \textsc{Ensemble} (0.751; $+0.111$ [$0.065$, $0.154$]). The decomposition explains why: adapter-only MC Dropout yields negligible mutual information (0.0001 nats), because dropout at $p_{\text{drop}}{=}0.05$ perturbs too little of the 4.38M-parameter adapter to vary the output, so its predictions barely differ from a single pass. The ensemble produces meaningful mutual information (0.057 nats) but still loses on error detection. A scalar temperature is the only add-on that helps, tightening calibration (Table~\ref{tab:calib}) without changing the ranking. For the configurations tested, the takeaway is to use the cheapest signal: one forward pass of softmax entropy, optionally temperature-scaled.

\subsection{Cross-Architecture Replication}\label{sec:res:cross}

The link holds on a different architecture. On PadChest, LLaVA-RAD LoRA predicts flips at AUROC 0.83 [0.799, 0.861] on its 732-item evaluation split (the pipeline holds out a random 15\% of the bank for temperature calibration; the 129 held-out items are excluded from scoring, not filtered on difficulty), consistent with MedGemma's 0.823 on the full 861-item bank, and is equally invariant to the flip definition (0.833 operator-preserving, 0.83 ex-negation). In-distribution on MIMIC (167 items, after the same holdout) it reaches 0.905. The base LLaVA-RAD model over-predicts ``yes'' on PadChest (89.6\% positive), which inflates its flip-prediction AUROC (0.95): a near-constant classifier rarely flips, and the few items it does flip are its least confident, so entropy separates them easily even though the classifier is clinically unhelpful. We therefore read the fine-tuned model as the cross-architecture check. The signal is consistent across two model families and two chest-radiograph datasets, and across five MedGemma seeds.

\section{Discussion and Conclusion}\label{sec:discussion}

\paragraph{One margin, two failures.}
Predictive entropy links error and paraphrase instability because in a binary answer head both are functions of the logit margin: near-boundary answers are the ones a rewrite can move. That relation is analytic; its measured strength is not, and here it is strong and stable: AUROC 0.82--0.83 across two chest-radiograph datasets, two architectures, and five seeds, and invariant to how a flip is defined. A system that already estimates uncertainty gets paraphrase screening for free from one threshold, with no operator or semantic filter at inference.

\paragraph{Deployment.}
One forward pass of softmax entropy covers calibration, error detection, and paraphrase screening. Because the fine-tuned model is accurate, the screen is not costly: at a 5\% risk budget it still answers 88\% of cases. Abstained cases need a defined escalation path --- the standard unassisted read, or a second reader --- so the screen adds triage without withdrawing care. A scalar temperature, fit once on held-out data, is a near-free calibration add-on; under cross-site shift the multi-pass methods we test (adapter dropout at $p_{\text{drop}}{=}0.05$, seed-only ensembling) give no further gain. Entropy thresholds should be recalibrated on each deployment site's validation set.

\paragraph{Limitations.}
We study binary presence VQA on chest radiographs, two datasets, and a single cross-site transition (MIMIC$\rightarrow$PadChest); other modalities, site pairs, and free-text generation, which would need semantic entropy~\cite{kuhn2023semantic}, are untested. We do not claim flip probability is monotone in entropy. The in-distribution evaluation ($n{=}196$) is question-disjoint from training but shares images with it, as the MIMIC-CXR splits are built per question; this can lift in-distribution accuracy more than the entropy-flip link, whose paraphrases are unseen, and PadChest carries the primary out-of-distribution evidence. Our multi-pass baselines are deliberately plain: MC Dropout perturbs only the 4.38M-parameter adapter at dropout 0.05, and the ensemble averages five seed-only adapters uniformly, so the single-pass verdict covers these configurations under this shift, not ensembling in general. Paraphrase equivalence rests on the bank's rule-based operator annotations and its blinded clinician audit of a stratified sample~\cite{sadanandan2026psfaudit}; only 28 of our 861 items carry an individually reviewed rewrite, so the audit validates how the bank was built rather than every rewrite we score. Independently written (non-LLM) paraphrases, member selection, weighting, joint calibration, weight-space methods (Laplace, SWAG, DDU), a second cross-site transition, and entailment-based semantic entropy remain future work.

\paragraph{Conclusion.}
Single-pass predictive entropy is a strong, stable, and cheap joint screen for unreliable and rephrase-sensitive answers in binary presence VQA on chest radiographs, on the two model families we test. It holds across our datasets, architectures, and seeds, needs no semantic filter, and under cross-site shift is not improved by the multi-pass variants we test.

\paragraph{Code and data availability.}
\begin{sloppypar}
Code and analysis-ready results are available at \url{https://github.com/thedatasense/predictive_entropy_unsure} (MIMIC-CXR needs credentialed PhysioNet access; PadChest is public).
\end{sloppypar}

\paragraph{Use of Large Language Models.}
The paraphrase bank we evaluate on, from the PSF-Med benchmark~\cite{sadanandan2026psfmed}, was generated and filtered by an LLM (\texttt{gpt-5-mini}); these paraphrases are a component of the benchmark under study, not a writing aid. Large language models also assisted with figure formatting and LaTeX layout. The authors designed the study, ran all experiments, and analyzed and validated all results.

\begin{credits}
\subsubsection{\ackname} Work conducted at the Secure and Assured Intelligent Learning (SAIL) Lab, University of New Haven. No specific external funding was received.

\subsubsection{\discintname} B.~Sadanandan is a research engineer at Medtronic Medical Surgical, CT, USA; this employment is unrelated to the present work. The authors cite related primary studies of their own~\cite{sadanandan2026psfmed,sadanandan2026psfaudit,sadanandan2026psflora}, published or under review elsewhere. No other competing interests are declared.
\end{credits}

\end{document}